\documentclass[10pt,twocolumn,letterpaper]{article}

\usepackage{iccv}
\usepackage{times}
\usepackage{epsfig}
\usepackage{graphicx}
\usepackage{amsmath}
\usepackage{amssymb}
\usepackage[caption=false]{subfig}

\usepackage[pagebackref=true,breaklinks=true,letterpaper=true,colorlinks,bookmarks=false]{hyperref}

\iccvfinalcopy % *** Uncomment this line for the final submission

\def\iccvPaperID{5961} % *** Enter the ICCV Paper ID here

\ificcvfinal\fi
\begin{document}

%%%%%%%%% TITLE
\title{SynFace: Face Recognition with Synthetic Data}

\author{Haibo Qiu$^{1}$\thanks{Equal contribution}, \ 
        Baosheng Yu$^{2}$\footnotemark[1], \ 
        Dihong Gong$^3$, \ 
        Zhifeng Li$^3$, \ 
        Wei Liu$^3$, \ 
        Dacheng Tao$^{1,2}$\\
$^1$ JD Explore Academy, China \quad
$^2$ The University of Sydney, Australia \\
$^3$ Tencent Data Platform, China\\
{\tt\small qiuhaibo1@jd.com, baosheng.yu@sydney.edu.au, gongdihong@gmail.com,} \\ {\tt\small michaelzfli@tencent.com, wl2223@columbia.edu, dacheng.tao@gmail.com}
% For a paper whose authors are all at the same institution,
% omit the following lines up until the closing ``}''.
% Additional authors and addresses can be added with ``\and'',
% just like the second author.
% To save space, use either the email address or home page, not both

}

\maketitle
% Remove page # from the first page of camera-ready.
% \ificcvfinal\fi
\ificcvfinal\thispagestyle{empty}\fi
%%%%%%%%% ABSTRACT
\begin{abstract}
With the recent success of deep neural networks, remarkable progress has been achieved on face recognition. However, collecting large-scale real-world training data for face recognition has turned out to be challenging, especially due to the label noise and privacy issues. Meanwhile, existing face recognition datasets are usually collected from web images, lacking detailed annotations on attributes (e.g., pose and expression), so the influences of different attributes on face recognition have been poorly investigated. In this paper, we address the above-mentioned issues in face recognition using synthetic face images, i.e., SynFace. Specifically, we first explore the performance gap between recent state-of-the-art face recognition models trained with synthetic and real face images. We then analyze the underlying causes behind the performance gap, e.g., the poor intra-class variations and the domain gap between synthetic and real face images. Inspired by this, we devise the SynFace with identity mixup (IM) and domain mixup (DM) to mitigate the above performance gap, demonstrating the great potentials of synthetic data for face recognition. Furthermore, with the controllable face synthesis model, we can easily manage different factors of synthetic face generation, including pose, expression, illumination, the number of identities, and samples per identity. Therefore, we also perform a systematically empirical analysis on synthetic face images to provide some insights on how to effectively utilize synthetic data for face recognition. Code is available at \url{https://github.com/haibo-qiu/SynFace}
\end{abstract}

%%%%%%%%% BODY TEXT

\section{Introduction}

\begin{figure}
    \centering
    \includegraphics[width=.99\linewidth]{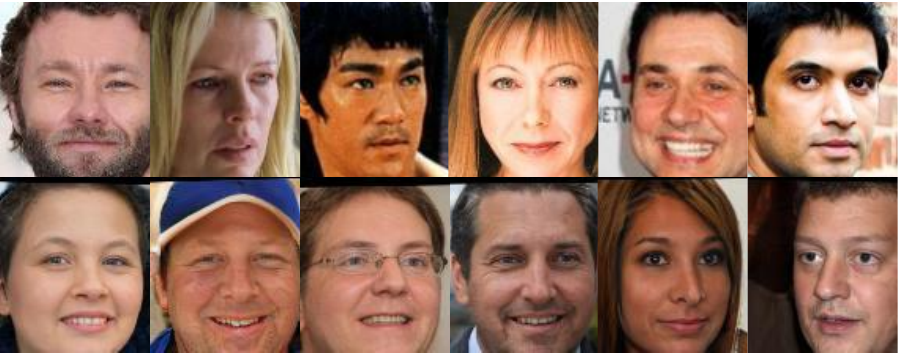}
    \caption{Examples of real/synthetic face images. The first row indicates real face images from CASIA-WebFace, and the second row shows synthetic face images generated by DiscoFaceGAN~\cite{deng2020disentangled} with the proposed identity mixup module.}
    \label{fig:face}
\end{figure}

In the last few years, face recognition has achieved extraordinary progress in a wide range of challenging problems including pose-robust face recognition~\cite{Cao2018, huang2000pose,yang2021larnet}, matching faces across ages~\cite{gong2013hidden,gong2015maximum,wang2019decorrelated,wang2018orthogonal}, across modalities~\cite{deng2019mutual,gong2017heterogeneous,gong2013multi,li2016mutual,li2014common}, and occlusions~\cite{qiu2021occ,song2019occlusion,zhang2007local}. Among these progresses, not only the very deep neural networks~\cite{he2016deep,huang2017densely,krizhevsky2017imagenet,simonyan2014very} and sophisticated design of loss functions~\cite{deng2019arcface,hoffer2015deep,liu2017sphereface,wang2018cosface,wen2016discriminative}, but also large-scale training datasets~\cite{guo2016ms,huang2008labeled,kemelmacher2016megaface} played important roles. However, it has turned out to be very difficult to further boost the performance of face recognition with the increasing number of training images collected from the Internet, especially due to the severe label noise and privacy issues~\cite{guo2016ms,wang2018devil,wang2019co}. For example, several large-scale face recognition datasets are struggling with the consent of all involved person/identities, or even have to close the access of face data from the website~\cite{guo2016ms}. Meanwhile, many face training datasets also suffer from the long-tailed problem, \ie, head classes with a large number of samples and tail classes with a few number of samples~\cite{liu2019large,ouyang2016factors,zhu2014capturing}. To utilize these datasets for face recognition, people need to carefully design the network architectures and/or loss functions to alleviate the degradation on model generalizability brought by the long-tailed problem. Furthermore, the above-mentioned issues also make it difficult for people to explore the influences of different attributes (\eg, expression, pose and illumination). 

To address the aforementioned issues, we explore the potentials of synthetic images for face recognition in this paper. Recently, face synthesis using GANs~\cite{goodfellow2014generative} and 3DMM~\cite{blanz1999morphable} have received increasing attention from the computer vision community, and existing methods usually focus on generating high-quality identity-preserving face images~\cite{bao2018towards,shen2018faceid,yin2017towards}. Some synthetic and real face images are demonstrated in Figure~\ref{fig:face}. However, the problem of face recognition using synthetic face images has not been well-investigated~\cite{kortylewski2019analyzing,trigueros2018generating}. Specifically, Trigueros \etal~\cite{trigueros2018generating} investigated the feasibility of data augmentation with photo-realistic synthetic images. Kortylewski \etal~\cite{kortylewski2019analyzing} further explored the pose-varying synthetic images to reduce the negative effects of dataset bias. Lately, disentangled face generation has become popular~\cite{deng2020disentangled}, which can provide the precise control of targeted face properties such as identity, pose, expression, and illumination, thus making it possible for us to systematically explore the impacts of facial properties on face recognition. Specifically, with a controllable face synthesis model, we are then capable of 1) collecting large-scale face images of non-existing identities without the risk of privacy issues; 2) exploring the impacts of different face dataset properties, such as the depth (the number of samples per identity) and the width (the number of identities); 3) analyzing the influences of different facial attributes (\eg, expression, pose, and illumination).

However, there is usually a significant performance gap between the models trained on synthetic and real face datasets. Through the empirical analysis, we find that 1) the poor intra-class variations in synthetic face images and 2) the domain gap between synthetic and real face datasets are the main reasons of the performance degradation. To address the above issues, we introduce identity mixup (IM) into the disentangled face generator to enlarge the intra-class variations of generated face images. Specifically, we use a convex combination of the coefficients from two different identities to form a new intermediate identity coefficient for synthetic face generation. Experimental results in Sec.~\ref{sec:experiments} show that the identity mixup significantly improves the performance of the model trained on synthetic face images. Furthermore, we observe a significant domain gap via cross-domain evaluation: 1) training on synthetic face images and testing on real face images; 2) training on real face images and testing on synthetic face images (see more details in Sec.~\ref{sec:explore}). Therefore, we further introduce the domain mixup (DM) to alleviate the domain gap, \ie, by using a convex combination of images from a large-scale synthetic dataset and a relatively small number of real face images during training. With the proposed identity mixup and domain mixup, we achieve a significant improvement over the vanilla SynFace, further pushing the boundary of face recognition performance using synthetic data. The main contributions of this paper are as follows:

\begin{itemize}
    \item We observe a performance gap between the models trained on real and synthetic face images, which can be effectively narrowed by 1) enlarging the intra-class variations via identity mixup; 2) leveraging a few real face images for domain adaption via domain mixup.
    \item We discuss the impacts of synthetic datasets with different properties for face recognition, \eg, depth (the number of samples per identity) and width (the number of identities), and reveal that the width plays a more important role.
    \item We analyze the influences of different facial attributes on face recognition (\eg, facial pose, expression, and illumination).
\end{itemize}

\section{Related Work}

\begin{figure*}[t]
    \centering
    \includegraphics[width=.99\textwidth]{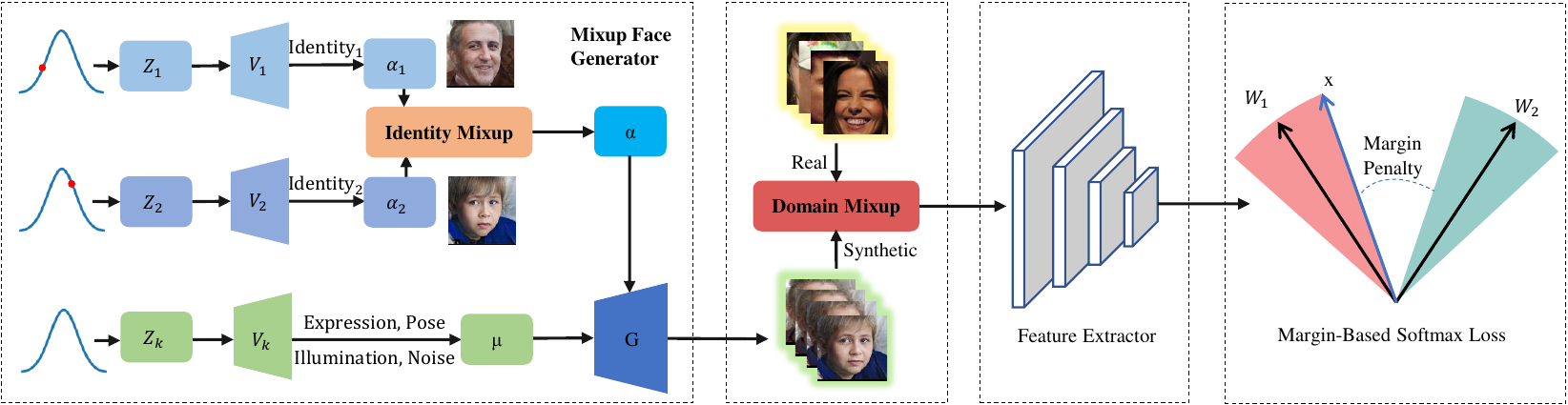}
    \caption{An overview of the proposed SynFace. Firstly, the identity mixup is introduced into DiscoFaceGAN~\cite{deng2020disentangled} to form the Mixup Face Generator, which can generate face images with different identities and their intermediate states. Next, the synthetic face images are cooperating with a few real face images via domain mixup to alleviate the domain gap. Then, the feature extractor takes the mixed face images as input and extracts the corresponding features. The extracted features are either utilized to calculate the margin-based softmax loss (where $W_1, W_2$ are the center weight vectors for two different classes and $x$ is the feature vector) for model training, or employed as the face representations to perform face identification and verification tasks.}
    \label{fig:pipeline}
\end{figure*}

We first briefly introduce visual tasks using synthetic data. Then recent face synthesis and recognition methods are reviewed. Lastly, we discuss the mixup and its variants to indicate their similarities/differences with the proposed identity mixup and domain mixup.

\textbf{Synthetic Data.} Synthetic data for computer vision tasks has been widely explored, \eg, crowd counting~\cite{wang2019learning}, vehicle re-identification~\cite{tang2019pamtri}, semantic segmentation~\cite{chen2019learning,sakaridis2018model,sankaranarayanan2018learning}, 3D face reconstruction~\cite{richardson20163d} and face recognition~\cite{kortylewski2019analyzing,trigueros2018generating}. According to the motivation, existing methods can be categorized into three groups: (1) It is time-consuming and expensive to collect and annotate large-scale training data~\cite{chen2019learning,richardson20163d,sakaridis2018model,sankaranarayanan2018learning}; (2) It can be used to further improve the model trained on a real dataset~\cite{kortylewski2019analyzing,trigueros2018generating}; (3) It can be used to systematically analyze the impacts of different dataset attributes~\cite{kortylewski2019analyzing}. Among these works, \cite{kortylewski2019analyzing} is the most related one to our work, while it only discusses the impacts of  different head poses. Apart from facial attributes (\eg, pose, expression, and illumination), we also explore the impacts of the width and the depth of training dataset. Furthermore, we introduce identity mixup (IM) and domain mixup (DM) to increase the intra-class variations and narrow down the domain gap, leading to a significant improvement.

\textbf{Face Synthesis.} With the great success of GANs~\cite{arjovsky2017wasserstein,Chen_2020_CVPR,goodfellow2014generative,mirza2014conditional,mroueh2017mcgan,qi2020loss,radford2015unsupervised}, face synthesis has received increasing attention and several methods have been proposed to generate identity-preserving face images~\cite{bao2018towards,shen2018faceid,yin2017towards}. Specifically, FF-GAN~\cite{yin2017towards} utilizes 3D priors (\eg, 3DMM~\cite{blanz1999morphable}) for high-quality face frontalization. Bao \etal \cite{bao2018towards} first disentangled identity/attributes from the face image, and then recombined different identities/attributes for identity-preserving face synthesis. FaceID-GAN~\cite{shen2018faceid} aims to generate identity-preserving faces by using a classifier (C) as the third player, competing with the generator (G) and cooperating with the discriminator (D). However, unlike exploring the identity-preserving property, generating face images from multiple disentangled latent spaces (\ie, different facial attributes) have not been well-investigated. Recently, DiscoFaceGAN~\cite{deng2020disentangled} introduces a novel disentangled learning scheme for face image generation via an imitative-contrastive paradigm using 3D priors. Thus, it further enables precise control of targeted face properties such as unknown identities, pose, expression, and illumination, yielding the flexible and high-quality face image generation.

\textbf{Deep Face Recognition.} Recent face recognition methods mainly focus on delivering novel loss functions for robust face recognition in the wild. The main idea  is to maximize the inter-class variations and minimize the intra-class variations. For example, 1) contrastive loss~\cite{chopra2005learning,hadsell2006dimensionality} and triplet loss~\cite{hoffer2015deep,yu2018correcting} are usually utilized to increase the Euclidean margin for better feature embedding; 2) center loss~\cite{wen2016discriminative} aims to learn a center for each identity and then minimizes the center-aware intra-class variations; 3) Large-margin softmax loss~\cite{liu2017sphereface,liu2016large} and its variants such as CosFace~\cite{wang2018cosface} and ArcFace~\cite{deng2019arcface} improve the feature discrimination by adding marginal constraints to each identity. 

\textbf{Mixup.} Mixup~\cite{zhang2017mixup} uses the convex combinations of two data samples as a new sample for training, regularizing deep neural networks to favor a simple linear behavior in-between training samples. Vanilla mixup is usually employed on image pixels, while the generated data samples are not consistent with the real images, \eg, a mixup of two face images in the pixel level does not always form a proper new face image. Inspired by this, we introduce identity mixup to face generator via the identity coefficients, where a convex combination of two identities forms a new identity in the disentangled latent space. With the proposed identity mixup, we are also able to generate high-fidelity face images correspondingly. Recently, several mixup variants have been proposed to perform feature-level interpolation~\cite{guo2019mixup,summers2019improved,takahashi2018ricap,verma2019manifold}, while \cite{xu2020adversarial} further leverages domain mixup to perform adversarial domain adaptation. Inspired by this, we perform domain adaption via domain mixup between real and synthetic face images, while the main difference is that \cite{xu2020adversarial} uses the mixup ratio to guide the model training, but we utilize the identity labels of both synthetic and real face images as the supervision for face recognition.

\section{Method}

In this section, we introduce face recognition with synthetic data, \ie, SynFace, and the overall pipeline is illustrated in Figure~\ref{fig:pipeline}. We first introduce deep face recognition using margin-based softmax loss functions. We then explore the performance gap between the models trained on synthetic and real datasets (SynFace and RealFace). Lastly, we introduce 1) identity mixup to enlarge the intra-class variations and 2) domain mixup to mitigate the domain gap between synthetic and real faces images.

\subsection{Deep Face Recognition}
\label{sec:loss}

With the great success of deep neural networks, deep learning-based embedding learning has become the mainstream technology for face recognition to maximize the inter-class variations and minimize the intra-class variations~\cite{chopra2005learning,hadsell2006dimensionality,hoffer2015deep,liu2016large}. Recently, margin-based softmax loss functions have been very popular in face recognition due to their simplicity and excellent performance, which explicitly explore the margin penalty between inter- and intra-class variations via a reformulation of softmax-based loss function~\cite{deng2019arcface,liu2017sphereface,wang2018cosface,yu2019deep}. Similar to~\cite{deng2019arcface}, we use a unified formulation for margin-based softmax loss functions as follows:
\begin{equation}
    \begin{split}
        & \mathcal{L}_{margin} =-\frac{1}{N}\sum_{i=1}^{N}\log\frac{e^{s \cdot \delta }}{e^{s \cdot \delta }+\sum_{j\neq  y_i}^{n}e^{s\cos\theta_{j}}},
%        \\
%        \\
%        & \wrt \quad \delta = \cos(m_1\theta_{y_i}+m_2) - m_3
    \end{split}
    \label{eq:margin}
\end{equation}
where $\delta = \cos(m_1\theta_{y_i}+m_2) - m_3$, $m_{1,2,3}$ are margins, $N$ is the number of training samples, $\theta_j$ indicates the angle between the weight $W_j$ and the feature $x_i$, $y_i$ represents the ground-truth class, and $s$ is the scale factor. Specifically, for SphereFace~\cite{liu2017sphereface}, ArcFace~\cite{deng2019arcface} and CosFace~\cite{wang2018cosface}, we have the coefficients $(m_1, 0, 0)$, $(0, m_2, 0)$, and $(0, 0, m_3)$, respectively, and we use ArcFace~\cite{deng2019arcface} as our baseline.

\subsection{SynFace \vs RealFace}
\label{sec:explore}

\begin{table}[ht]
\begin{center}
\begin{tabular}{|c|c|c|c|}
\hline
Method & Training Dataset & LFW & Syn-LFW \\
\hline
\hline
RealFace & CASIA-WebFace & 99.18 & 98.85\\
SynFace & Syn\_10K\_50 & 88.98 & 99.98\\
\hline
\end{tabular}
\end{center}
\caption{The cross-domain evaluation of SynFace and RealFace using the metric of face verification accuracy ($\%$).}
\label{tab:cross-domain}
\end{table}

To explore the performance gap between SynFace and RealFace, as well as the underlying causes, we perform experiments on real-world face datasets and synthetic face datasets generated by DiscoFaceGAN~\cite{deng2020disentangled}. Specifically, for real-world face datasets, we use CASIA-WebFace~\cite{yi2014learning} for training and LFW~\cite{huang2008labeled} for testing. For the fair comparison, we generate the synthetic version of the LFW dataset, Syn-LFW, using the same parameters (the number of samples, the number of identities, distributions of expression, pose, and illumination). For synthetic training data, we generate $10K$ different identities with $50$ samples per identity to form a comparable training dataset to CASIA-WebFace (containing 494,414 images from 10,575 subjects) and we refer to it as Syn\_10K\_50. More details of synthetic dataset construction can be found in Sec.~\ref{sec:dataset}. With both synthetic and real face images, we then perform the cross-domain evaluation as follows. We train two face recognition models on CASIA-WebFace and Syn\_10K\_50, and test them on LFW and Syn-LFW, respectively. As shown in Table~\ref{tab:cross-domain}, there is a clear performance gap ($88.98\%$ \vs $99.18\%$) when testing on LFW, while SynFace outperforms RealFace on Syn-LFW ($99.98\%$  \vs  $98.85\%$). These observations suggest that the domain gap between synthetic and real face images contributes to the performance gap between SynFace and RealFace.

\begin{figure}
    \centering 
    \includegraphics[width=0.9\linewidth]{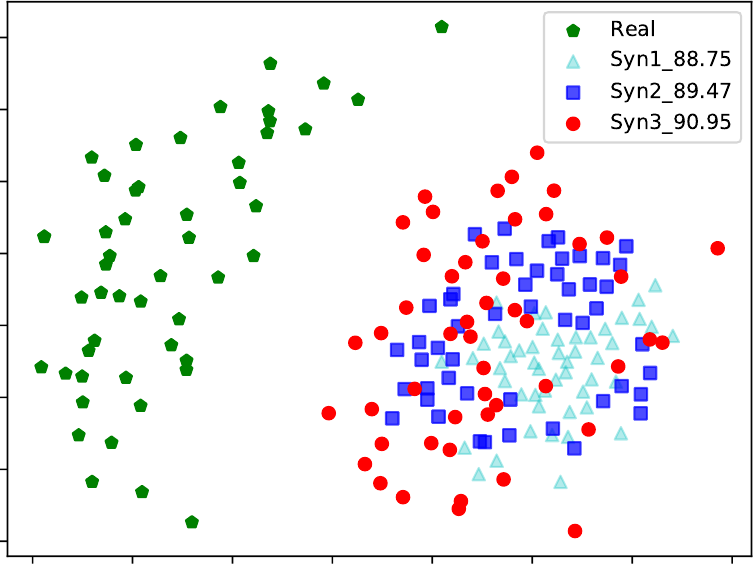}
    \caption{Visualization of the feature distributions (using MDS~\cite{borg2005modern}) for the samples from three different synthetic datasets (Syn1, Syn2 and Syn3) and CASIA-WebFace, which are illustrated by the cyan triangles, blue square, red circle and green pentagon, respectively. Note that the intra-class variations of Syn1, Syn2 and Syn3 are increasing, which lead to the consistent improvements on accuracy ($88.75\% \rightarrow 89.47\% \rightarrow 90.95\%$). Best viewed in color.}
    \label{fig:diversity}
\end{figure}

We compare the face images between Syn\_10K\_50 and CASIA-WebFace, and find that the synthetic face images usually lack the intra-class variations, which may be one of the reasons for the performance degradation (please refer to the supplementary materials for more illuminations). Furthermore, we also visualize the distributions of feature embeddings by using multidimensional scaling (MDS~\cite{borg2005modern}) to convert the 512-dimensional feature vector into 2D space.  As shown in Figure~\ref{fig:diversity}, we randomly select 50 samples from two different classes of Syn\_10K\_50 and CASIA-WebFace, respectively. In particular, we observe that the cyan triangles have a much more compact distribution than the green pentagons, suggesting the poor intra-class variations in Syn\_10K\_50. 

\subsection{SynFace with Identity Mixup}
\label{sec:generator}

To increase the intra-class variations of synthetic face images, we incorporate the identity mixup into DiscoFaceGAN~\cite{deng2020disentangled} to form a new face generator for face recognition, \ie, the Mixup Face Generator, which is capable of generating different identities and their intermediate states. In this subsection, we first briefly discuss the mechanism of DiscoFaceGAN, and we then introduce how to incorporate the proposed identity mixup into the face generator.

\textbf{Face Generator}. DiscoFaceGAN~\cite{deng2020disentangled} can provide the disentangled, precisely-controllable latent representations for the identity of non-existing people, expression, pose, and illumination to generated face images. Specifically, it generates realistic face images $x$ from random noise $z$, which consists of five independent variables $z_i\in\mathbb{R}^{N_i}$ and each of them follows a standard normal distribution. The above five independent variables indicate independent factors for face generation: identity, expression, illumination, pose, and random noise accounting for other properties such as the background. Let $\lambda \doteq[\alpha, \beta,\gamma,\theta]$ denote the latent factors, where $\alpha, \beta,\gamma $ and $\theta$ indicate the identity, expression, illumination, and pose coefficient, respectively. Four simple VAEs~\cite{dai2019diagnosing} of $\alpha$, $\beta$, $\gamma$ and $\theta$ are then trained for $z$-space to $\lambda$-space mapping, which enables training the generator to imitate the rendered faces from 3DMM~\cite{blanz1999morphable}. The pipeline of generating a face image is to 1) first randomly sample latent variables from the standard normal distribution, 2) then feed them into the trained VAEs to obtain $\alpha, \beta,\gamma $ and $\theta$ coefficients, and 3) the corresponding face image is synthesized by the generator using these coefficients.

\begin{figure}
    \centering
    \includegraphics[width=0.9\linewidth]{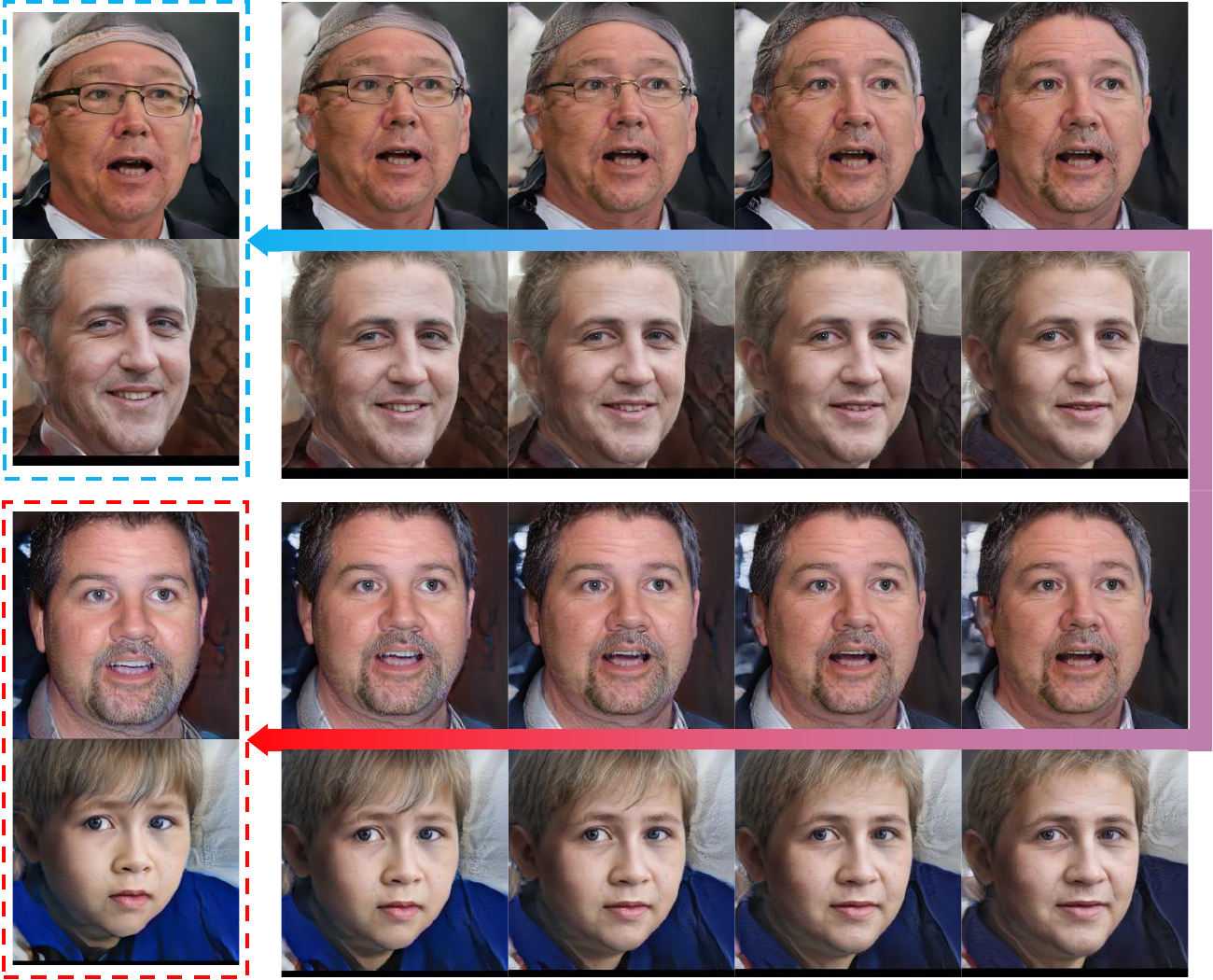}
    \caption{ Examples of an identity gradually and smoothly varying to another identity as the weighted ratio $\varphi$ varies from 0 to 1.}
    \label{fig:interpretation}
\end{figure}

\textbf{Identity Mixup (IM)}. Inspired by the reenactment of face images~\cite{zhang2020freenet}, we propose to enlarge the intra-class variations by interpolating two different identities as a new intermediate one with changing the label correspondingly. Recalling that the coefficient $\alpha$ controls the identity characteristic, we thus interpolate two different identity coefficients to generate a new intermediate identity coefficient. Mathematically, it can be formulated as follows:

\begin{equation}
    \begin{split}
        \alpha &= \varphi \cdot \alpha_{1} + (1 - \varphi) \cdot \alpha_{2},
        \\
        \eta &= \varphi \cdot \eta_1 + (1 - \varphi) \cdot \eta_2,
    \end{split}
    \label{eq:idmixup}
\end{equation}
where $\alpha_1$, $\alpha_2$ are two random identity coefficients from $\lambda$-space, and $\eta_1$, $\eta_2$ are the corresponding class labels. Note that the weighted ratio $\varphi$ is randomly sampled from the linear space which varies from $0.0$ to $1.0$ with interval being 0.05 (\ie, $np.linspace(0.0, 1.0, 21)$). Comparing to the vanilla mixup~\cite{zhang2017mixup} which is employed at the pixel level, the proposed mixup is operating on the identity coefficient latent space, denoted as identity mixup (IM), which enlarges the intra-class variations by linearly interpolating different identities, forming the Mixup Face Generator. However, both of them can regularize the model to favor the simple linear behavior in-between training samples.

As illustrated in Figure~\ref{fig:pipeline}, the pipeline of Mixup Face Generator is first randomly sampling two different identity latent variables from the standard normal distribution, and then feeding them to the trained VAEs to obtain $\alpha_1, \alpha_2$ coefficients. The mixed identity coefficient $\alpha$ is obtained by identity mixup with $\alpha_1, \alpha_2$ according to Eq.~\eqref{eq:idmixup}, the corresponding face image is finally synthesized by the generator with $\alpha, \mu$ coefficients (where $\mu \doteq[\beta,\gamma,\theta]$). We also visualize two groups of identity interpolation with identity mixup in Figure~\ref{fig:interpretation}. As we can see, one identity gradually and smoothly transforms to another identity as the weighted ratio $\varphi$ varies from 0 to 1. Besides, from Figure~\ref{fig:interpretation}, it is obvious that the face images generated with intermediate identity coefficients are also high-quality.

To evaluate the identity mixup for enlarging the intra-class variations, as illustrated in Figure~\ref{fig:diversity}, we visualize the feature embedding distributions of the same class in three synthetic datasets (containing $5K$ different identities with $50$ samples per identity) with different levels of identity mixup (IM) by using multidimensional scaling (MDS~\cite{borg2005modern}). Note that Syn1, Syn2 and Syn3 represent the weighted ratio $\varphi$ is 1.0 (\ie, no IM), 0.8 and randomly sampled from the linear space which varies from $0.6$ to $1.0$ with the interval being 0.05 (\ie, $np.linspace(0.6, 1.0, 11)$). It is clear that the cyan triangles (Syn1) have the smallest variations , while the red circles (Syn3) have the largest one and the blue squares (Syn2) are in the middle position. Accordingly, the accuracy is in an increasing trend (\ie, $88.75\% \rightarrow 89.47\% \rightarrow 90.95\%$). Besides, $88.98\%$ (as in Table~\ref{tab:cross-domain}) is boosted to $91.97\%$ (as in Table~\ref{tab:mix}) after utilizing identity mixup. In particular, when the baseline is weaker, the improvement brought by identity mixup is larger, which are shown in Table~\ref{fig:depth_width} and Figure~\ref{fig:fatcors}.

\begin{figure}[t]
    \centering
    \includegraphics[width=.9\linewidth]{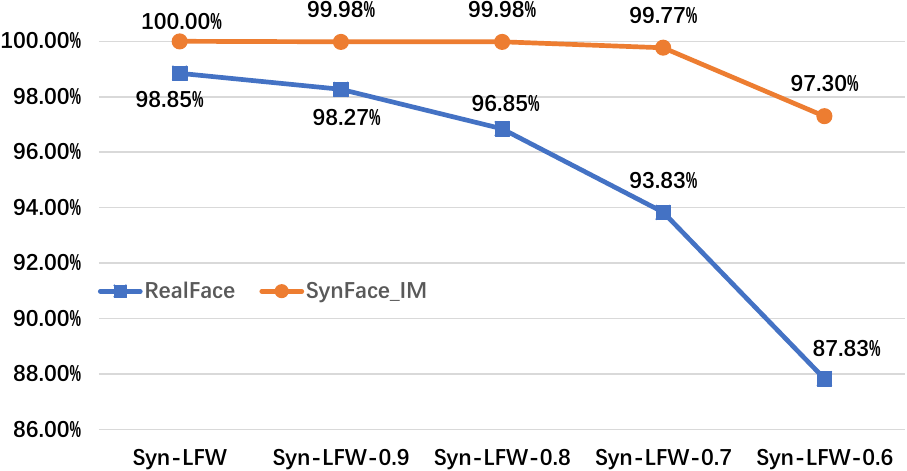}
    \caption{Face verification accuracy comparison between RealFace and SynFace\_IM (\ie, SynFace with Identity Mixup) on five different synthetic testing datasets. Syn-LFW is the synthetic version of the LFW dataset, while Syn-LFW-R (with R $\in [0.6, 0.7, 0.8, 0.9]$) indicates introducing identity mixup with ratio R into Syn-LFW.}
    \label{fig:domain_diff}
\end{figure}

In addition to utilizing identity mixup in the training process, we also make an attempt of employing identity mixup on the synthetic testing dataset to evaluate the model's robustness on the identity coefficient noises. Specifically, both RealFace (trained on CASIA-WebFace) and SynFace\_IM (trained on Syn\_10K\_50 with identity mixup) are evaluated on five different synthetic testing datasets, as illustrated in Figure~\ref{fig:domain_diff}. Note that Syn-LFW is the synthetic version of the LFW dataset, while Syn-LFW-R (with R $\in [0.6, 0.7, 0.8, 0.9]$) indicates employing the identity mixup with the weighted ratio R during the generation of Syn-LFW. Specifically, we mix the primary class with a random secondary class using the ratio R according to Eq.~\eqref{eq:idmixup}, but we keep the original label unchanged. Apparently, when R is smaller (\ie, the weight of the primary class is smaller), the corresponding testing dataset is more difficult to recognize because the secondary class impacts the identity information more heavily. 

From the results of Figure~\ref{fig:domain_diff}, we can find that our SynFace\_IM achieves nearly perfect accuracy when R is larger than 0.6, and also obtains an impressive $97.30\%$ result which remarkably outperforms the $87.83\%$ accuracy by RealFace when R is $0.6$. On the other hand, the accuracy of RealFace drops significantly on Syn-LFW-R when R becomes small, which suggests that the domain gap between real and synthetic face data is still large even after employing the identity mixup. Another interesting conclusion is that the current state-of-the-art face recognition model (\ie, RealFace) cannot handle the identity mixup attack. In other words, if a face image is mixup with another identity, the model cannot recognize it well. However, the proposed SynFace with identity mixup can nearly keep the accuracy under the identity mixup attack. We prefer to explore how to make the RealFace handle such an attack in future work.

\subsection{SynFace with Domain Mixup}

The lack of intra-class variation is an observable cause of the domain gap between synthetic and real faces, and SynFace can be significantly improved by the proposed identity mixup. To further narrow the performance gap between SynFace and RealFace, we introduce the domain mixup as a general domain adaptation method to alleviate the domain gap for face recognition. Specifically, we utilize large-scale synthetic face images with a small number of real-world face images with labels as the training data. When training, we perform mixup within a mini-batch of synthetic images and a mini-batch of real images, where the labels changed accordingly as the supervision. Mathematically, the domain mixup can be formulated as follows:  
\begin{equation}
    \begin{split}
        X &= \psi \cdot X_{S} + (1 - \psi) \cdot X_{R},
        \\
        Y &= \psi \cdot Y_{S} + (1 - \psi) \cdot Y_{R},
    \end{split}
    \label{eq:dmixup}
\end{equation}
where $X_{S}, X_{R}$ indicate the synthetic and real face images, respectively, and $Y_{S}, Y_{R}$ indicate their corresponding labels. Note that $\psi$ is the mixup ratio which is randomly sampled from the linear space distribution from $0.0$ to $1.0$ with the interval being 0.05 (\ie, $np.linspace(0.0, 1.0, 21)$). For the large-scale synthetic data, we synthesize the Syn\_10K\_50 dataset that has $10K$ different identities with $50$ samples per identity. For a small set of real-world data, we utilize the first $2K$ identities of CASIA-WebFace. The experimental results are shown in Table~\ref{tab:mix}. Specifically, the first row, Syn\_10K\_50, indicating the baseline method without using any real face images, achieves the accuracy $91.97\%$ using identity mixup. ``Real\_N\_S'' means the use of only real images, $N$ identities with $S$ samples per identity during training, while ``Mix\_N\_S'' indicates a mixture of $N$ real identities with $S$ samples per identity with Syn\_10K\_50 during training. Both identity mixup and domain mixup are employed on all the `Mix\_N\_S'' datasets. As demonstrated in Table~\ref{tab:mix}, domain mixup brings a significant and consistent improvement over the baseline methods under different settings. For example, Mix\_2K\_10 obtains $95.78\%$ accuracy, which significantly surpasses $91.97\%$ achieved by Syn\_10K\_50 and $91.22\%$ achieved by Real\_2K\_10. We conjecture that mixup with the real images can bring the real-world appearance attributes (\eg, blur and illumination) to synthetic images, which alleviate the domain gap. If we continue to increase the number of real images for training, \eg, Mix\_2K\_20, the performance can be further boosted from $95.78\%$ to $97.65\%$.

\begin{table}[t]
\begin{center}
\begin{tabular}{|l|c|c|c|}
\hline
Method & R\_ID & Samples per R\_ID & Accuracy \\
\hline
Syn\_10K\_50 & 0 & 0 & 91.97 \\
\hline
Real\_1K\_10 & 1K & 10 & 87.50 \\
Mix\_1K\_10 & 1K & 10 & \textbf{92.28} \\
\hline
Real\_1K\_20 & 1K & 20 & 92.53 \\
Mix\_1K\_20 & 1K & 20 & \textbf{95.05} \\
\hline
Real\_2K\_10 & 2K & 10 & 91.22 \\
Mix\_2K\_10 & 2K & 10 & \textbf{95.78} \\
\hline
\end{tabular}
\end{center}
\caption{Face verification accuracies ($\%$) of models trained on synthetic, real and mixed datasets on LFW. R\_ID means the number of real identities.}
\label{tab:mix}
\end{table}

\section{Experiments}
\label{sec:experiments}

With the introduced Mixup Face Generator, we are able to generate large-scale face images with controllable facial attributes, including the identity, pose, expression,  illumination, and other dataset characteristics such as the depth and the width. In this section, we perform empirical analysis using synthetic face images. Specifically, we first introduce the datasets (Sec.~\ref{sec:dataset}) and the implementation details (Sec.~\ref{sec:implement}). Then the long-tailed problem is mitigated by employing the balanced synthetic face dataset and identity mixup (Sec.~\ref{sec:long}). Lastly, we analyze the impacts of depth, width (Sec.~\ref{sec:depth_width}), and different facial attributes (Sec.~\ref{sec:facial}). 

\subsection{Datasets}
\label{sec:dataset}
\textbf{Real Datasets.} We employ the CASIA-WebFace~\cite{yi2014learning} and LFW~\cite{huang2008labeled} for training and testing, respectively. The CASIA-WebFace dataset contains around 500,000 web images, \ie, 494,414 images from 10,575 subjects. The LFW dataset is a widely-used benchmark for face verification, which contains 13,233 face images from 5,749 identities. Following the protocol in \cite{deng2019arcface}, we report the verification accuracy on 6,000 testing image pairs.

\textbf{Synthetic Datasets.} We first generate a synthetic version of LFW, in which all synthetic face images share the same properties with LFW images, \eg, expression, illumination, and pose. Specifically, for each image in LFW, we first use the 3D face reconstruction network in~\cite{deng2019accurate} to obtain the attribute coefficients $\mu \doteq[\beta,\gamma,\theta]$, which indicate the expression, illumination and pose coefficient, respectively. We then adopt the DiscoFaceGAN~\cite{deng2020disentangled} to generate the face images according to these attribute coefficients with a random identity coefficient. Finally, we obtain a new dataset and refer to it as Syn-LFW, which has the same statistics as LFW with unknown identities (non-existing people). For synthetic training dataset (\eg, Syn\_10K\_50), we construct it by randomly sampling latent variables from the standard normal distribution for identity, expression, pose and illumination coefficients, respectively, which leads to the same person with different expressions, poses and illuminations in the same class. Note that the identities of Syn-LFW do not have the overlap with any synthetic training datasets.

\subsection{Implementation Details}
\label{sec:implement}

We use the MTCNN~\cite{zhang2016joint} to detect face bounding boxes and five facial landmarks (two eyes, nose and two mouth corners). All face images are then cropped, aligned (similarity transformation), and resized to $112 \times 96$ pixel as illustrated in Figure~\ref{fig:face}. Similar to~\cite{deng2019arcface,wang2018cosface}, we normalize the pixel values (in $[0, 255]$) in RGB images to $[-1.0, 1.0]$ for training and testing. To balance the trade-off between the performance and computational complexity, we adopt the variant of ResNet~\cite{he2016deep}, LResNet50E-IR, as our backbone framework, which is devised in ArcFace~\cite{deng2019arcface}. All models are implemented with PyTorch~\cite{NEURIPS2019_9015} and trained from scratch using Eight NVIDIA Tesla V100 GPUs. We use the additive angular margin loss defined in Eq.~\eqref{eq:margin}, \ie, with $(m_1, m_2, m_3) = (0, 0.5, 0)$ and $s=30$. If not mentioned, we always set the batch size to 512. We use SGD with a momentum of $0.9$ and a weight decay of $0.0005$. The learning rate starts from 0.1, and is divided by 10 at the 24, 30 and 36 epochs, with 40 epochs in total.

\subsection{Long-tailed Face Recognition}
\label{sec:long}

\begin{figure}
\centering
\includegraphics[width=.9\linewidth]{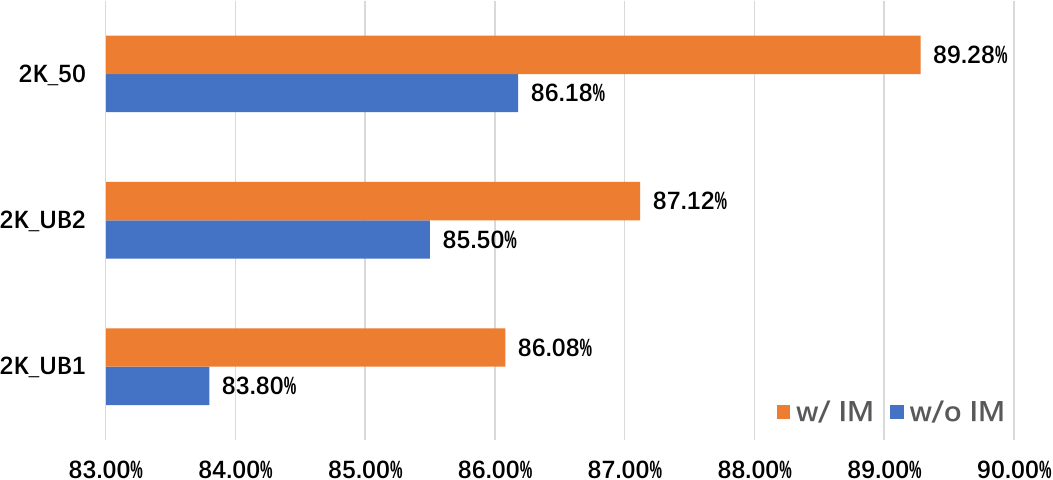}
\caption{Face verification accuracies ($\%$) on LFW using the training datasets with decreasing imbalance, \ie, ``2K\_UB1'', ``2K\_UB2'', and ``2K\_50'', where we assign $N$ defined in Eq.(4) as $[2, 2, 6, 40, 200]$, $[4, 16, 30, 80, 120]$, and $[50, 50, 50, 50, 50]$, respectively. w/ IM and w/o IM indicate whether identity mixup (IM) is used during training.}
\label{fig:long_tail}
\end{figure}

\textbf{Experimental Setup.}
To explore the long-tailed problem, we construct multiple synthetic datasets with the purpose that each dataset has the same number of identities ($2K$) and total images ($100K$) but different degrees of unbalance. Face images are generated using the equation:
\begin{equation}
    \begin{split}
        N &= [N_1, N_2, N_3, N_4, N_5],\\
        ID &= [400, 400, 400, 400, 400],
    \end{split}
    \label{eq:construct}
\end{equation}
where $ID$ indicates the number of identities in each of the five groups, and $N$ means the number of samples of the five groups. For example, if $N=[30, 40, 50, 60, 70]$, the corresponding synthetic dataset has $400$ identities with $30$ samples per identity, and the rest $1600$ identities with $40, 50, 60, 70$ samples per identity, respectively. We construct three different synthetic datasets by assigning $N$ to be $[2, 2, 6, 40, 200]$, $[4, 16, 30, 80, 120]$ and $[50, 50, 50, 50, 50]$, which are denoted as ``2K\_UB1'', ``2K\_UB2'' and ``2K\_50'', respectively. The detailed construction process can be found in Sec.~\ref{sec:dataset}. Note that all the three datasets have average $50$ samples per identity, while the first two have unbalanced distributions with the standard deviations $76.35$ and $43.52$, and the last one is the perfectly balanced dataset.

\textbf{Empirical Analysis.}
We train face recognition models on the above three different synthetic datasets and the experimental results are illustrated in Figure~\ref{fig:long_tail}. We see that the model trained on the ``2K\_UB1'' achieves the worst performance ($83.80\%$), suggesting that the long-tailed problem or the unbalanced distribution leads to the degradation of the model performance. Comparing with the models trained on ``2K\_UB1'' and ``2K\_UB2'', we discover that decreasing the degree of unbalance leads to the improvement on the performance. Finally, when the model is trained on ``2K\_50'', \ie, the perfectly balanced dataset, the accuracy is significantly improved to $86.18\%$. Therefore, with balanced synthetic data, the long-tailed problem can be intrinsically avoided. Besides, introducing the identity mixup for training can consistently and significantly improve the performance over all the settings.

\subsection{Effectiveness of ``Depth'' and ``Width''}
\label{sec:depth_width}
\textbf{Experimental Setup.}
We synthesize multiple face datasets with different width (the number of identities) and depth (the number of samples per identity). Let ``$N\_S$'' denote the synthetic dataset containing $N$ identities with $S$ samples per identity, \eg, $1K\_50$ indicates the dataset having $1K$ different identities and $50$ samples per identity. Obviously, $N$ and $S$ represent the dataset's width and depth. The details of dataset construction can be found in Sec.~\ref{sec:dataset}.

\textbf{Empirical Analysis.}
We train the same face recognition model on these synthetic datasets, and the experimental results (both w/wo identity mixup) are shown in Table~\ref{fig:depth_width}. Firstly, we analyze the influence of the width of the dataset by comparing the results of $(a),~(b),~(c),~(i)$. From $(a)$ to $(c)$, we see that the accuracy dramatically increases from $83.85\%$ to $88.75\%$. However, the improvement is marginal from $(c)$ to $(i)$, which implies that the synthetic data may suffer from the lack of inter-class variations. Observing the results of $(d),~(e),~(f),~(g),~(h),~(i)$, we conclude that the accuracy significantly increases with the increasing of dataset depth, but it is quickly saturated when the depth is larger than $20$, which is in line with the observation on real data made by Schroff~\etal~\cite{schroff2015facenet}. Lastly, we see that  $(a)$ and $(e)$ have the same number of total images (50K), while $(a)$ outperforms $(e)$ with a large margin, \ie, $4.37\%$, which reveals that the dataset width plays as the more important role than the dataset depth in term of the final face recognition accuracy. Similar observation can be found by comparing $(b)$ and $(f)$. Importantly, employing the identity mixup (IM) for training consistently improves the performance over all the datasets, which confirms the effectiveness of IM. The best accuracy $91.97\%$ brought by IM significantly outperforms the original $88.98\%$.

\begin{table}[t]
\begin{center}
\begin{tabular}{|l|c|c|c|c|}
\hline
Method & ID  & Samples & LFW & LFW(w/ IM)\\
\hline\hline
$(a)$ 1K\_50 & 1K & 50 & 83.85 & \textbf{87.53} \\
$(b)$ 2K\_50 & 2K & 50 & 86.18 & \textbf{89.28} \\
$(c)$ 5K\_50 & 5K & 50 & 88.75 & \textbf{90.95} \\
\hline
$(d)$ 10K\_2 & 10K & 2 & 78.85 & \textbf{80.30} \\
$(e)$ 10K\_5 & 10K & 5 & 88.22 & \textbf{88.32} \\
$(f)$ 10K\_10 & 10K & 10 & 89.48 & \textbf{90.28} \\
$(g)$ 10K\_20 & 10K & 20 & 89.90 & \textbf{90.87} \\
$(h)$ 10K\_30 & 10K & 30 & 89.73 & \textbf{91.17} \\
$(i)$ 10K\_50 & 10K & 50 & 88.98 & \textbf{91.97} \\
\hline
\end{tabular}
\end{center}
\caption{Face verification accuracies ($\%$) on LFW~\cite{yi2014learning}. ``$N\_S$'' implies that the corresponding dataset has $N$ identities with $S$ samples per identity, \ie, $N$ and $S$ indicate the width and depth. LFW (w/ IM) means employing the identity mixup (IM) for training.}
\label{fig:depth_width}
\end{table}

\subsection{Impacts of Different Facial Attributes}
\label{sec:facial}

\textbf{Experimental Setup.} We explore the impacts of different facial attributes for face recognition (\ie, expression, pose and illumination) by controlling face generation process. We construct four synthetic datasets that have $5K$ identities and $50$ samples per identity. The difference between the four datasets is the distribution of different facial attributes. Specifically, the first dataset is referred to as ``Non'', since it fixes all the facial attributes. The rest three datasets are referred to as ``Expression'', ``Pose'', and ``Illumination'', respectively, which indicates the only changed attribute while keeping other attributes unchanged. 

\begin{figure}
    \centering
    \includegraphics[width=.9\linewidth]{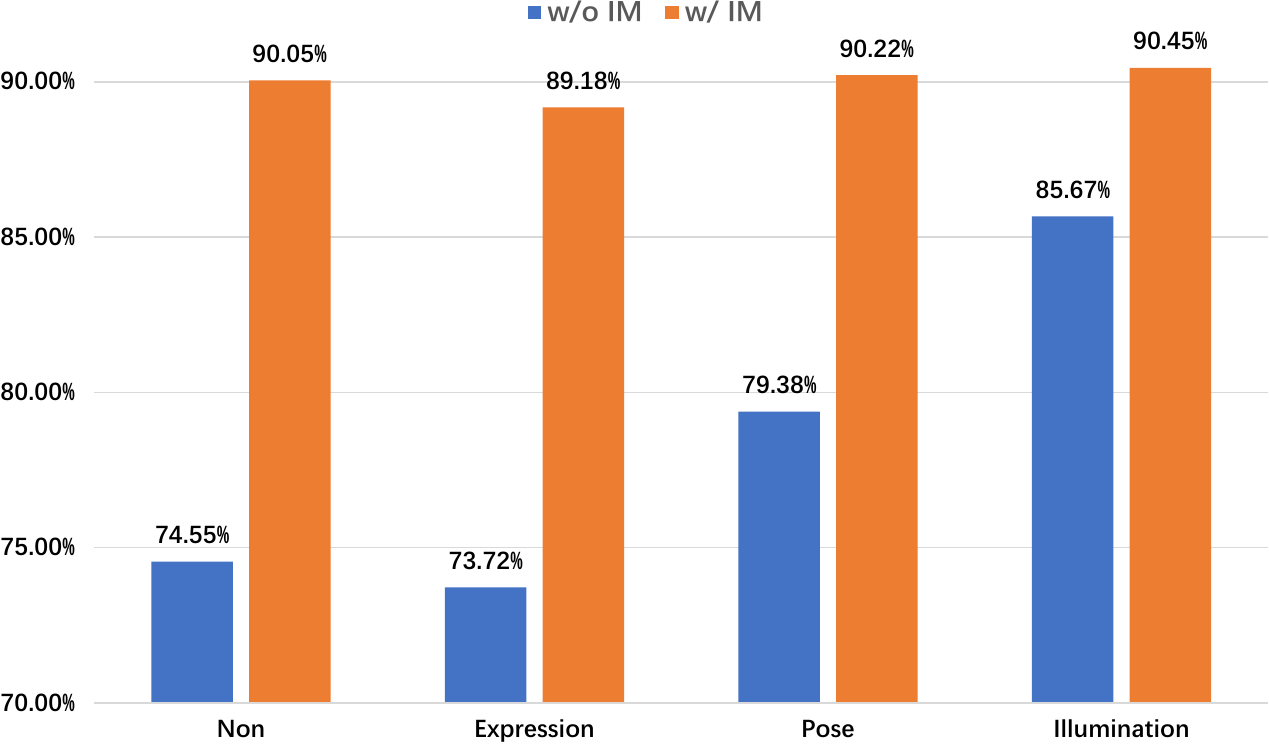}
    \caption{Face verification accuracies ($\%$) on LFW using the training datasets with variations in different facial attributes. Specifically, ``Expression'', ``Pose'', and ``Illumination'' indicate that we separately introduce variations in expression, pose, and illumination while keeping the other attributes unchanged. w/ IM and w/o IM indicate whether identity mixup (IM) is used during training.}
    \label{fig:fatcors}
\end{figure}

\textbf{Empirical Analysis.}
As shown in Figure~\ref{fig:fatcors}, ``Non'' and ``Expression'' achieve the worst two performances $74.55\% $ and $ 73.72\%$. Specifically, we find that ``Expression'' is limited to poor diversity, \ie, the generated face images mainly have the expression of ``smiling" (see more demo images in the supplementary materials). Hence, there is basically only one valid sample per identity for ''Non'' and ''Expression'', causing the poor performances. Experimental results on ``Pose'' and ``Illumination'' demonstrate significant improvements over ``Non'', possibly due to their more diverse distributions and the testing dataset (\ie, LFW) also has similar pose and illumination. Lastly, we find that all of four settings are significantly improved with the proposed identity mixup, especially for ``Non''. A possible reason is that identity mixup can be regarded as a strong data augmentation method for face recognition, reducing the influences of different facial attributes on the final recognition accuracy. 

\section{Conclusion}

In this paper, we explored the potentials of synthetic data for face recognition, \ie, SynFace. We performed a systematically empirical analysis and provided novel insights on how to efficiently utilize synthetic face images for face recognition: 1) enlarging the intra-class variations of synthetic data consistently improves the performance, which can be achieved by the proposed identity mixup; 2) both the depth and width of the training synthetic dataset have significant influences on the performance, while the saturation first appears on the depth dimension, \ie, increasing the number of identities (width) is more important; 3) the impacts of different attributes vary from pose, illumination and expression, \ie, changing pose and illumination brings significant improvements, while generated face images suffer from a poor diversity on expression; 4) a small subset of real-world face images can greatly boost the performance of SynFace via the proposed domain mixup. 

\section*{Acknowledgement}
Dr. Baosheng Yu is supported by ARC project FL-170100117.

{\small
\bibliographystyle{ieee_fullname}
\bibliography{iccv2021}
}

\begin{figure*}[]
    \centering
    \subfloat[CASIA-WebFace]{\includegraphics[width=.48\linewidth]{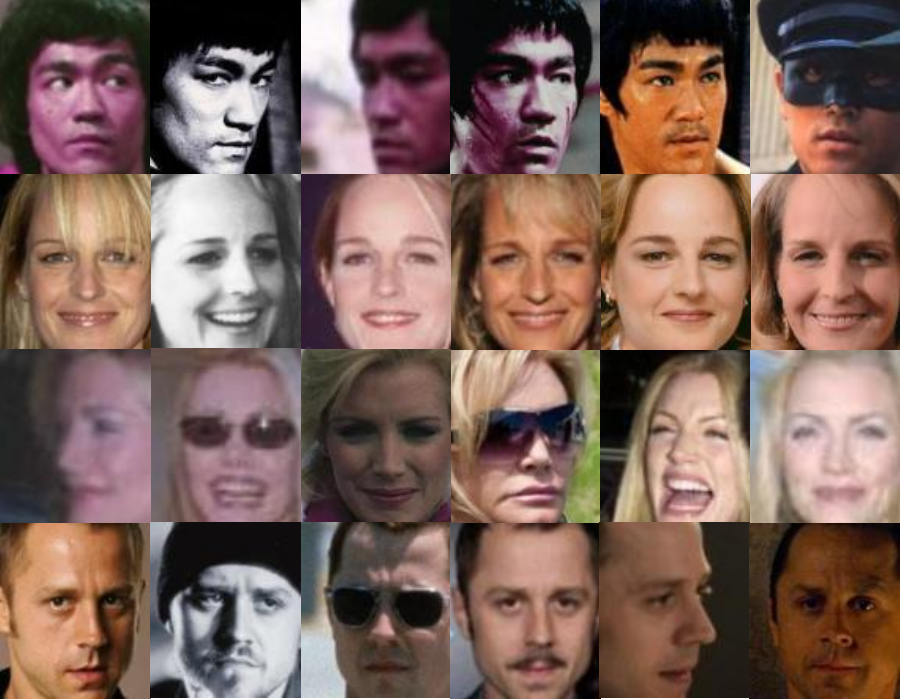} \label{fig:sup1}}\quad
    \subfloat[Syn\_10K\_50]{\includegraphics[width=.48\linewidth]{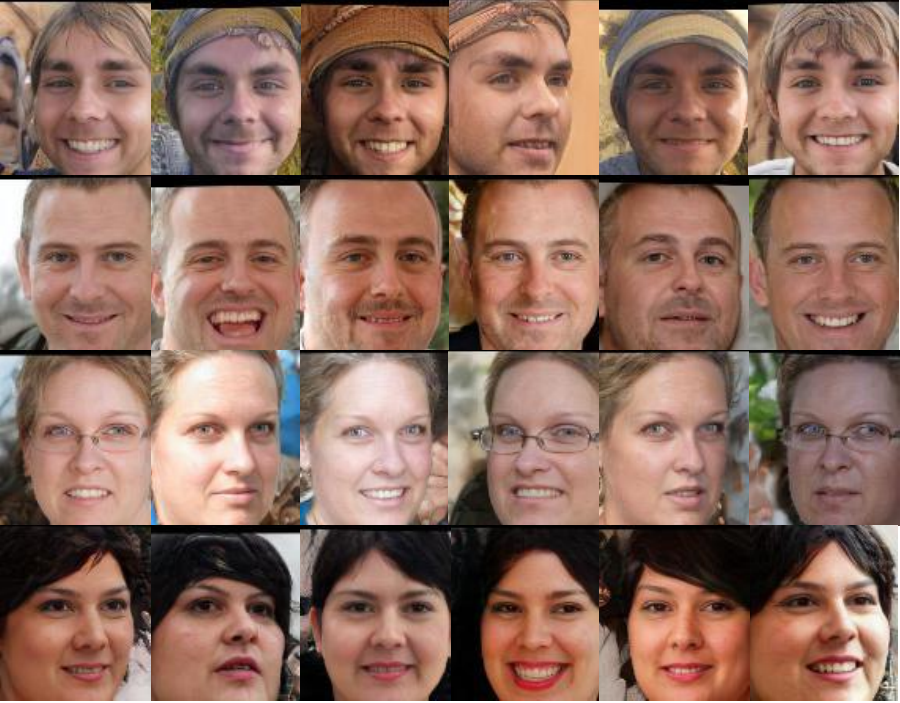} \label{fig:sup2}}
    \caption{Comparison of real and synthetic face images. Each row indicates the same person with different face images. Obviously, comparing the real-world dataset, the synthetic dataset significantly lacks of the intra-class variations.}
    \label{fig:sup}
\end{figure*}

\newpage
\appendix
In this appendix, we illustrate plenty of face images (from both Syn\_10K\_50 and CASIA-WebFace) to further demonstrate our observations: 1) the synthetic dataset usually lacks of intra-class variations which significantly degrades the performance (Appendix.~\ref{sec:var}), and 2) the generated face images have limited diversity on facial expressions which are mainly ``smiling'' with slight differences (Appendix.~\ref{sec:div}).  

\section{Intra-class Variations}
\label{sec:var}
Recalling that there is a clear performance gap ($88.98\%$ \vs $99.18\%$) on LFW~\cite{huang2008labeled} between SynFace and RealFace. We notice that the fundamental purpose of face synthesis model (\eg, DiscoFaceGAN~\cite{deng2020disentangled}) is to generate high-quality and clean face images, while the face recognition model is usually required to recognize those face images in the wild (\eg, LFW~\cite{huang2008labeled}) with complex conditions. Therefore, this kind of domain gap leads to the model trained on synthetic data intrinsically lacking well generalization ability. 

Then we explore the potential factors which are responsible for the simplicity of Syn\_10K\_50. Figure~\ref{fig:sup} demonstrates multiple face images of different people from both CASIA-WebFace (Figure~\ref{fig:sup1}) and Syn\_10K\_50 (Figure~\ref{fig:sup2}), in which face images of one row belong to the same person. As we can observe, the variations of real face images are clearly larger than the synthetic images. For example, comparing to the synthetic face images, the real face images in the wild usually have the large motion blur and illumination variations. If we augment the synthetic face images with the ColorJitter transformation in PyTorch and MotionBlur from Albumentations~\cite{info11020125} for training, the face recognition performance is boosted from $88.98\%$ to $91.23\%$. Hence, we conclude that the lack of intra-class variations by synthetic dataset leads to its simplicity which significantly degrades the face recognition performance.

\section{Expression Diversity}
\label{sec:div}

\begin{figure*}
    \centering
    \includegraphics[width=.72\textwidth]{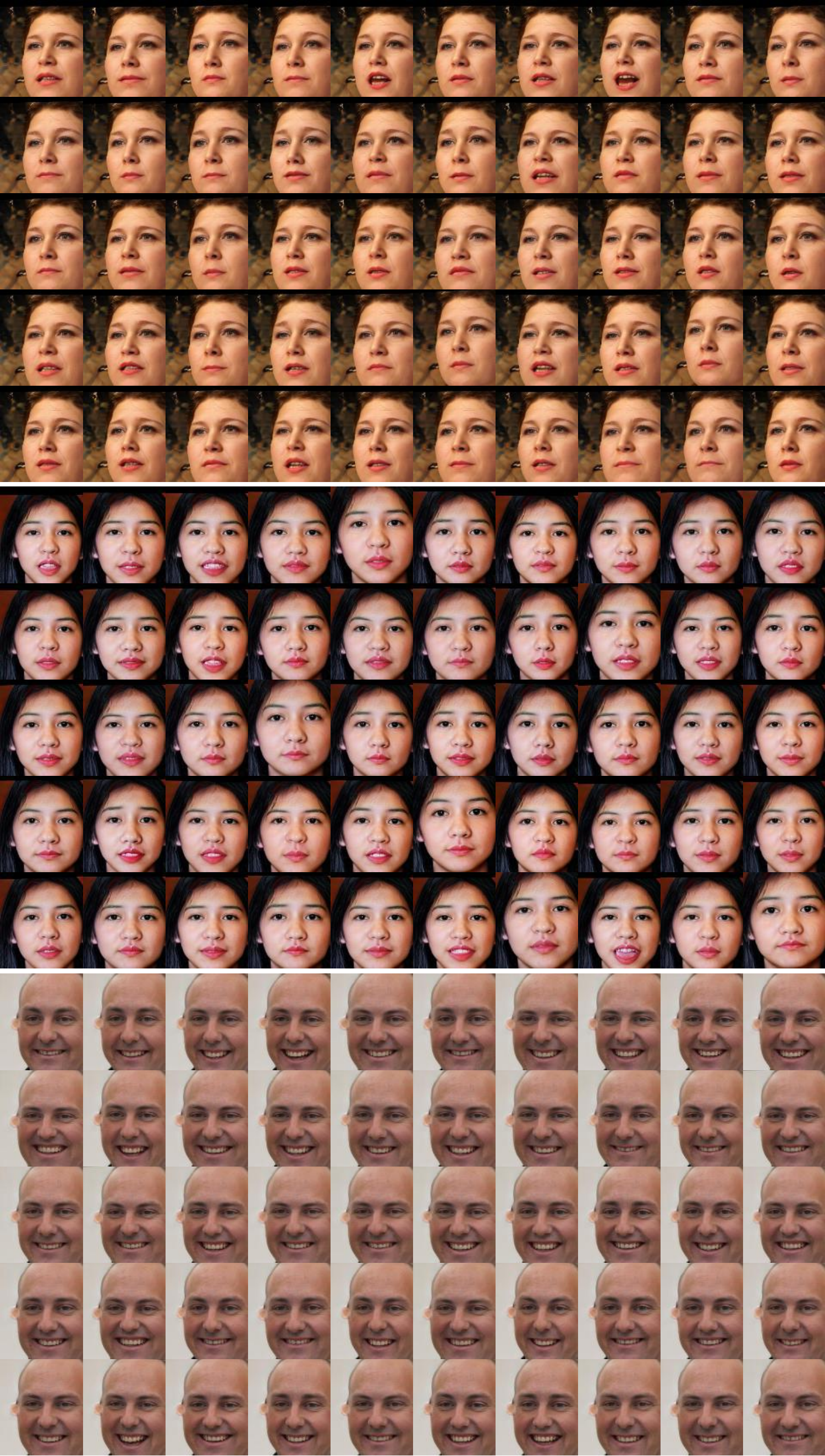}
    \caption{Visualizations of all the samples from three different classes. The generated expressions of face images are mainly ``smiling" despite of slight differences, which reveals the limited expression diversity of ``Expression''.}
    \label{fig:exp}
\end{figure*}

We randomly select three classes from the ``Expression'' dataset (which means only varying the facial expression of face images while fixing the other attributes) and visualize all the samples ($50$ images per identity) in Figure~\ref{fig:exp}. Apparently, the differences of images inside the same class are marginal and only reflected by the mouth variations, which reveal the limited expression diversity of ``Expression'' that is responsible for the worst performance. We conjecture that the 3D priors from 3DMM~\cite{blanz1999morphable} and the training images from web lack of the expression variations, which result in the limited expression diversity of synthetic face images generated by DiscoFaceFAN~\cite{deng2020disentangled}.

\end{document}